\documentclass{article}

\usepackage{microtype}
\usepackage{graphicx}
\usepackage{subcaption}
\usepackage{booktabs} 
\usepackage{multirow}

\usepackage{hyperref}

\usepackage[preprint]{icml2026}

\usepackage{amsmath}
\usepackage{amssymb}
\usepackage{mathtools}
\usepackage{amsthm}
\usepackage{booktabs}
\usepackage{enumitem}
\usepackage[dvipsnames, table]{xcolor}
\usepackage{wrapfig}
\definecolor{mygray}{rgb}{0.5,0.5,0.5}

\usepackage[capitalize,noabbrev]{cleveref}

\theoremstyle{plain}

\theoremstyle{definition}

\theoremstyle{remark}

\newcommand{\ourmethod}[1]{\textsc{MoCo}}

\usepackage[textsize=tiny]{todonotes}

\icmltitlerunning{MoCo: A One-Stop Shop for Model Collaboration Research}

\begin{document}

\twocolumn[
  \icmltitle{
  \raisebox{-0.3\height}{\includegraphics[height=3em]{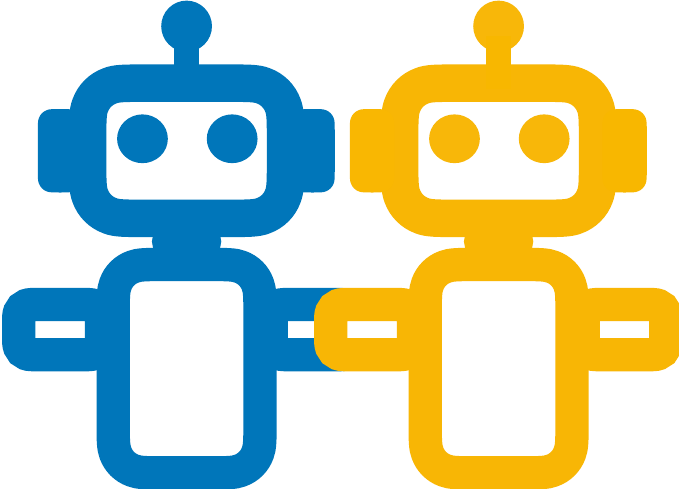}} \
  \ourmethod{}: A One-Stop Shop for Model Collaboration Research}



  \icmlsetsymbol{equal}{*}

  \begin{icmlauthorlist}
    \icmlauthor{Shangbin Feng}{equal,uw}
    \icmlauthor{Yuyang Bai}{equal,tamu}
    \icmlauthor{Ziyuan Yang}{equal,uw} \\
    \icmlauthor{Yike Wang}{uw}
    \icmlauthor{Zhaoxuan Tan}{nd}
    \icmlauthor{Jiajie Yan}{}
    \icmlauthor{Zhenyu Lei}{virginia}
    \icmlauthor{Wenxuan Ding}{nyu}
    \icmlauthor{Weijia Shi}{uw}
    \icmlauthor{Haojin Wang}{uiuc}
    \icmlauthor{Zhenting Qi}{harvard}
    \icmlauthor{Yuru Jiang}{uw}
    \icmlauthor{Heng Wang}{uiuc}
    \icmlauthor{Chengsong Huang}{uw2}
    \icmlauthor{Yu Fei}{uci}
    \icmlauthor{Jihan Yao}{uw} \\
    \icmlauthor{Yilun Du}{harvard}
    \icmlauthor{Luke Zettlemoyer}{uw}
    \icmlauthor{Yejin Choi}{stanford}
    \icmlauthor{Yulia Tsvetkov}{uw}
  \end{icmlauthorlist}

  \icmlaffiliation{uw}{University of Washington}
  \icmlaffiliation{tamu}{Texas A\&M University}
  \icmlaffiliation{nd}{University of Notre Dame}
  \icmlaffiliation{virginia}{University of Virginia}
  \icmlaffiliation{nyu}{New York University}
  \icmlaffiliation{uiuc}{University of Illinois Urbana-Champaign}
  \icmlaffiliation{harvard}{Harvard University}
  \icmlaffiliation{uw2}{Washington University in St. Louis}
  \icmlaffiliation{uci}{University of California Irvine}
  \icmlaffiliation{stanford}{Stanford University}

  \icmlcorrespondingauthor{Shangbin Feng}{shangbin@cs.washington.edu}


  \vskip 0.3in
]



\printAffiliationsAndNotice{}  

\vspace*{-10pt}
\begin{abstract}
  Advancing beyond single monolithic language models (LMs), recent research increasingly recognizes the importance of \emph{model collaboration}, where multiple LMs collaborate, compose, and complement each other. Existing research on this topic has mostly been disparate and disconnected, from different research communities, and lacks rigorous comparison. To consolidate existing research and establish model collaboration as a school of thought, we present \ourmethod{}: a one-stop Python library of executing, benchmarking, and comparing model collaboration algorithms at scale. \ourmethod{} features 26 model collaboration methods, spanning diverse levels of cross-model information exchange such as routing, text, logit, and model parameters.
  \ourmethod{} integrates 25 evaluation datasets spanning reasoning, QA, code, safety, and more, while users could flexibly bring their own data. Extensive experiments with \ourmethod{} demonstrate that most collaboration strategies outperform models without collaboration in 61.0\% of (model, data) settings on average, with the most effective methods outperforming by up to 25.8\%. We further analyze the scaling of model collaboration strategies, the training/inference efficiency of diverse methods, highlight that the collaborative system solves problems where single LMs struggle, and discuss future work in model collaboration, all made possible by \ourmethod{}. We envision \ourmethod{} as a valuable toolkit to facilitate and turbocharge the quest for an open, modular, decentralized, and collaborative AI future.\footnotemark[2]\footnotetext[2]{\href{https://github.com/BunsenFeng/model_collaboration}{https://github.com/BunsenFeng/model\_collaboration}}
\end{abstract}

\section{Introduction}
Language models (LMs) are increasingly not used in isolation but in \emph{collaboration}: multiple LMs discuss \citep{feng2024don}, debate \citep{du2023improving}, divide and conquer \citep{yu2025netsafe} to solve complex problems; multiple LMs form routing systems to select the best model for each user query \citep{ongroutellm, fenggraphrouter}; multiple LMs interact and exchange information in the logits \citep{liutuning} and model parameter \citep{yadavsurvey} space for collaborative decoding, generation, and deriving new models.

\begin{figure}
    \centering
    \includegraphics[width=0.8\linewidth]{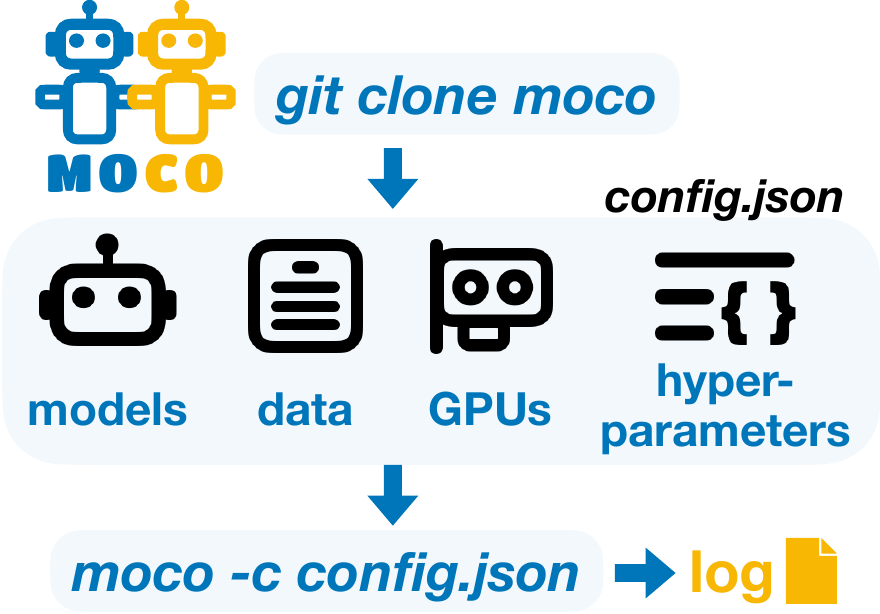}
    \caption{\ourmethod{} is a comprehensive library for model collaboration research. Download \ourmethod{}, write a config file specifying model collaboration setups (models, data, hardware, etc.), execute and compare diverse model collaboration algorithms with \ourmethod{}.}
    \vspace*{-10pt}
    \label{fig:overview}
\end{figure}

These efforts, despite being previously disconnected, unrelated, and underappreciated only as ad-hoc solutions to a narrow problem, together demonstrate the promise of \emph{model collaboration} \citep{feng2025one}: multiple language models, trained by different people, on different data, and thus possessing diverse skills and strengths, collaborate, compose, and complement each other. Model collaboration has the unique potential to unlock a collaborative and decentralized AI future, featuring modular and compositional AI systems built from the bottom up with everyone everywhere's models and contribution.

However, existing research on this topic are mostly disconnected and lack rigorous comparison. To consolidate existing research progress, evaluate diverse methods, motivate future work, and facilitate model collaboration's potential as compositional AI systems built by the many, we propose \ourmethod{}: a one-stop Python library and framework to build, execute, and compare diverse model collaboration systems:
\begin{itemize}[leftmargin=*]
    \item \ourmethod{} features a wide range of 26 model collaboration algorithms, spanning four levels of collaboration defined by the level of information exchange: API-level (e.g., routing \citep{ongroutellm} and switching \citep{feng2025don, huang2026relayllm}), text-level (e.g., debate \citep{du2023improving} and cooperate \citep{yu2025netsafe}), logit-level (e.g., collective decoding \citep{liutuning}), and weight-level (e.g., merging \citep{yadavsurvey} and parameter-space search \citep{feng2025model}).
    \item \ourmethod{} provides flexible implementations of model collaboration strategies, supporting their execution and evaluation with any amount of any LMs with any hardware setting (e.g., any amount of GPUs), democratizing model collaboration research for small models and less compute.
    \item \ourmethod{} comes with 25 built-in evaluation datasets (and growing), spanning reasoning, math, QA, knowledge, science, instruction following, safety, coding, computational social science, and more. Users could also flexibly bring their own prompts and datasets for evaluation and comparing model collaboration algorithms.
    \item Like model collaboration, we envision \ourmethod{} as a collaborative initiative: we provide detailed documentation and annotated code templates so everyone everywhere could contribute their model collaboration research to \ourmethod{}. We commit to providing continuous support for external contributors, even after publication.
\end{itemize}

Extensive experiments with \ourmethod{} demonstrate that model collaboration is a promising path towards modular and compositional AI systems. Model collaboration outperforms individual models in 61.0\% of cases across diverse (model, data) settings, with the most successful algorithms outperforming in almost every evaluation domain by up to 25.8\%. These results also enable reflection on existing methods and progress: text-level and weight-level collaboration generally work best across the board, reasoning tasks could be sensitive to model choice, and model collaboration algorithms benefit most from the diversity of language models. We further analyze the potential of scaling up these model collaboration systems, the training/inference efficiency of diverse methods, quantitative evidence where collaboration solves problems where individual models struggle, and motivate future work. We envision \ourmethod{} as a valuable framework to spearhead a new generation of modular, bottom-up, and compositional AI systems.

\section{\ourmethod{}}
\label{sec:method}
We introduce \ourmethod{}, a comprehensive Python library for model collaboration research. \ourmethod{} supports 26 diverse approaches spanning four levels of collaboration, incorporates 25 datasets and benchmarks for evaluation, and features great flexibility and extensibility.

\subsection{Methods}
We categorize the 26 model collaboration algorithms in \ourmethod{} into four collaboration levels, depending on the level of information exchange across LLMs.

\paragraph{API-level collaboration} approaches aim to select the best LLM in the pool for response generation, through paradigms such as routing, cascading, and switching.

\emph{Method \#1: Prompt Routing} Given an instruction, we prompt an LLM to select the best-fitting model in the pool based on their model descriptions provided by the user.

\emph{Method \#2: Nudging} A base LLM generates responses guided by one or more nudging LLMs: if the base LLM is uncertain about the next token (token probability lower than a threshold), then the nudging models generate these tokens (often stylistic/discourse markers) to guide the base model’s generation. \citep{fei2024nudging}

\emph{Method \#3: Switch Generation} Given a pool of LLMs, we train a selector LM to govern how multiple LMs take turns to generate token patches as part of the full response. \citep{feng2025don}

\emph{Method \#4: Trained Router} Given a pool of LLMs, we train a routing LM by evaluating models on the dev set, identifying the best model for each data point, and supervised fine-tuning the routing LM to select the best model. At inference-time, the trained router conducts routing and the selected models generate. \citep{ongroutellm}

\emph{Method \#5: Graph Router} Similar to Method \#4, but we employ a graph neural network operating on a task-query-model graph as the routing mechanism. \citep{fenggraphrouter}

\emph{Method \#6: Cascade} Given an ordered list of language models, we let each model generate first and defer to the next model in line if the current model is uncertain, in terms of token probabilities. \citep{chenfrugalgpt, guptalanguage}

\emph{Method \#7: Co-LLM} We train a small deferral model for a pair of LLMs to decide when a model should defer generation to another model. During inference, the deferral model and two LLMs are jointly employed to collaborative generate responses. \citep{shen2024learning}

\emph{Method \#8: Mentor Collab} Given a generator model and a mentor model, we stop the generator model at random token positions, inspect whether the next predicted token differs between the two models. If yes, the generator model or an additionally trained classifier decides which model to generate the immediate following text patch.

\paragraph{Text-level collaboration} approaches feature exchanges of generated texts among models, through paradigms such as ``debate'', ``feedback'', and ``discuss''.

\emph{Method \#9: Multiagent Debate} Given a pool of LLMs, each model first independently generates an answer, then refine their answer based on the answers of other LLMs. Repeats for a few iterations and an LLM summarizes the final responses. \citep{du2023improving}

\emph{Method \#10: Multiagent Feedback} Each model first independently generates an answer, then generates feedback for the answers of other models, and refines their own answer based on received feedback. Repeats for a few iterations and an LLM summarizes the final responses. \citep{feng2024don}

\emph{Method \#11: LLM Blender} Each model first generates an asnwer, then a ranker LLM, optionally trained with pairwise preferences on the dev set, conducts re-ranking of the responses. The top-k ranked responses are merged with a fuser LLM, optionally trained with gold answers on the dev set, to derive a final answer. \citep{jiang2023llm}

\emph{Method \#12: Knowledge Card} Given a question, each LLM generates a paragraph of related knowledge and information. An LLM then answers the question based on the aggregated knowledge of all LLMs. \citep{fengknowledge}

\emph{Method \#13: Majority Vote} Each model generates an answer, then a majority/plurality vote is employed to select the final answer. Only works for objective tasks with a definitive answer.

\emph{Method \#14: Heterogeneous Swarms} Multiple LLMs form a directed acyclic graph to collaboratively generate responses, where one LLM's output becomes part of another LLM input based on directed edges. The graph structure is optimized with particle swarm optimization on the dev set. \citep{feng2025heterogeneous}

\emph{Method \#15: Multiagent Finetuning} Each LLM first generates an initial response, then perform majority vote to get a consensus, and build generation and critic agent fine-tuning datasets for adaptation. Repeat for a few iterations for training. At inferece-time, the finetuned agents perform debate and the final answer is decided by majority vote. \citep{subramaniammultiagent}

\emph{Method \#16: Structured Interaction} Multiple LLMs interact and update their responses based on a specified graph structure, where each model receives the responses from models within its 1-hop neighborhood to update their answer. \citep{yu2025netsafe}

\emph{Method \#17: BBMAS} Multiple LLMs collaborate through a shared blackboard to iteratively solve problems. Specifically, models/agents take turns to contribute to the blackboard through a set of five actions. In the end, all models vote on the best final conclusion.

\emph{Method \#18: Sparta Alignment} Multiple LLMs collectievly self-align through competition and mutual evaluation. Specifically, two models are sampled to compete in fulfilling an instruction, while other LLMs judge the contest. The winning model gains in reputation and vice versa, which effects how much say does it have in evaluating other LLMs. We then perform preference optimization on the collected preferences, where the winning response is preferred over the losing response. \citep{jiang2025sparta}

\emph{Method \#19: AggLM} Each model first generates an response, we then train an aggregator model with reinforcement learning to aggregate their responses. Specifically, we employ GRPO \citep{shao2024deepseekmath} with verifiable rewards and balances training on ``hard'' examples (where majority voting fails) and ``easy'' examples (where majority voting succeeds) to learn both minority-answer recovery and reliable aggregation. \citep{zhao2025majority}

\paragraph{Logit-level collaboration} approaches feature employing and transforming the token probability distributions among multiple LLMs for collaboration.

\emph{Method \#20: Logit Fusion} We average the next-token probabilities across multiple LLMs and decode from the joint distribution for text generation. These LLMs need to share the same tokenization/vocabulary.

\emph{Method \#21: Logit Contrastive} We first evaluate the pool of LLMs on the dev set of the dataset. We then retain the top-k and bottom-k models, and decode with joint distribution $p_1 + \alpha (p_1 + \cdots + p_k - p_{n-k+1} - \cdots - p_{n})$, where $p_i$ denotes the token probabilities of the $i$-th ranked model and $\alpha$ is a hyperparameter. \citep{liutuning}

\paragraph{Weight-level collaboration} approaches feature arithmetic and merging in the model parameter space for collaboration. These methods often require the LLM pool to share the same architecture.

\emph{Method \#22: Greedy Soup} We first evaluate the pool of LLMs on the dev set and sort them in descending performance. Starting from the best model, we iteratively add one model at a time to the soup, the parameter averaging of all selected models, and retains it if it improves performance on the dev set. \citep{wortsman2022model}

\emph{Method \#23: Dare Ties} Models in the pool are merged with the DARE random pruning \citep{yu2024language} and the TIES sign consensus of parameter values \citep{yadav2023ties}. We incorporate the MergeKit implementation \citep{goddard-etal-2024-arcees} and gratefully acknowledge their valuable contribution.

\emph{Method \#24: Model Swarms} Multiple LLMs collaboratively search in the model weight space to find better parameter values based on dev set performance. The search is instanstiated with particle swarm optimization. \citep{feng2025model}

\emph{Method \#25: LoraHub} We use gradient-free optimization to learn the best scalar weights to linearly compose multiple language models, specifically LoRA adapters. \citep{huanglorahub}

\emph{Method \#26: ExPO} We employ model \textbf{ex}tra\textbf{po}lation in the parameter space over the pool of LLMs. Specifically, we first evaluate all LLMs on the dev set and rank by performance. We then merge the top-k models and bottom-k models, and extrapolate with $\mathbf{x}_{\textit{expo}} = \mathbf{x}_{\textit{top-k}} + \alpha (\mathbf{x}_{\textit{top-k}} - \mathbf{x}_{\textit{bottom-k}})$. \citep{zheng2024weak}

Please note that \ourmethod{} does not aim to be a reproducibility study: we adapt the core ideas behind related papers and employ what works flexibly. For a better understanding of any of the incorporated methods, please refer to the software repository for details. While \ourmethod{} provides a comprehensive slate of diverse model collaboration algorithms, it is not an exhaustive list: we encourage readers who work on model collaboration to get in touch and incorporate their method in \ourmethod{}.

\subsection{Evaluation}

To facilitate evaluations and fair comparisons, \ourmethod{} comes with 25 (and growing) evaluation datasets/benchmarks built-in, spanning diverse model capabilities.

\begin{itemize}[leftmargin=*]
    \item General-purpose QA: AGIEval \citep{zhong2024agieval}, ARC-challenge \citep{clark2018think}, MMLU-redux \citep{gema2025we}, GPQA \citep{rein2024gpqa}
    \item Math: GSM8k \citep{cobbe2021training}, MATH \citep{hendrycks2020measuring}
    \item Reasoning: BigBench-hard \citep{suzgun2023challenging}, TableMWP \citep{ludynamic}, TheoremQA \citep{chen2023theoremqa}
    \item Knowledge and factuality: WikiDYK \citep{zhang2025bidirectional}, PopQA \citep{mallen2023not}
    \item Human diversity: BLEND \citep{myung2024blend}, CultureBench \citep{chiu2024culturalbench}
    \item Science: Sciencemeter \citep{wang2025sciencemeter}, Sciriff \citep{wadden2025sciriff}
    \item Safety: TruthfulQA \citep{lin2022truthfulqa}, CocoNot \citep{brahman2024art}
    \item Coding: mbpp \citep{austin2021program}, HumanEval \citep{chen2021codex}
    \item Medical: MedQA \citep{jin2021disease, li2024mediq}, MedMCQA \citep{pal2022medmcqa}, PubMedQA \citep{jin2019pubmedqa}
    \item Instruction following: AlpacaEval \citep{dubois2023alpacafarm}, Wildchat \citep{zhaowildchat}, human interest \citep{feng2025model}
\end{itemize}
We by default downsample to 1k for both the dev and test sets, if the original dataset is large. These tasks and datasets provide a comprehensive test bed for diverse model collaboration algorithms in \ourmethod{}, while users could flexibly bring their own data/benchmarks for evaluation with \ourmethod{}.

\subsection{Design principles of \ourmethod{}}
\begin{itemize}[leftmargin=*]
    \item \textbf{Flexibility}: We provide flexible implementations for diverse model collaboration strategies, so that they could be executed and evaluated with any amount of any LLMs with any hardware setting (e.g., any amount of common GPUs). The commitment to democratize model collaboration research is core to our mission in \ourmethod{}, where small models and less compute are adequately supported.
    \item \textbf{Extensibility}: \ourmethod{} will not stop at 26 methods and 25 datasets. New methods from new/existing research could be flexibly added to \ourmethod{}: we provide blank code templates to guide contributors, and we commit to providing continuous support for external contributors, even after publication. Users could also fleixbly bring their own data for generation and evaluation, or contributing evaluations important to them as part of \ourmethod{}.
    \item \textbf{Foundational}: In addition to being an artifact building on top of a lot of amazing research, we envision that \ourmethod{} also servs as a solid foundation and opens up sweeping new research avenues about model collaboration, compositional AI, collaborative development, and more. How do we incorporate the models/contributions across diverse parties to jointly build an AI system? What is the scalability of diverse model collaboration approaches in terms of the number of models and model diversity? How would malicious models impact the performance/integrity of different collaboration systems? How do we increase the efficiency and reconcile the strengths/weaknesses of diverse collaboration algorithms? These critical research questions are all made possible to pursue based on the unified \ourmethod{} framework and infrastructure.
\end{itemize}

\begin{table*}[t]
\centering
\scriptsize
\setlength{\tabcolsep}{4pt}
\renewcommand{\arraystretch}{1.2}
\caption{Performance of \textcolor{NavyBlue}{API-level}, \textcolor{Dandelion}{text-level}, \textcolor{Maroon}{logit-level}, and \textcolor{OliveGreen}{weight-level} model collaboration algorithms over two model pool settings and six evaluation domains. Improvements over the best single model without collaboration are highlighted in \colorbox{orange!15}{orange}. IF stands for instruction following: we normalize its scores with min-max standardization to 0-1 and calculate the macro-average across evaluation domains for the ``avg.'' column. Best in \textbf{bold} and second-best in \underline{underline}. * indicates that these methods operate with objective tasks: the open-ended generation CocoNot dataset is not included in their performance on the safety domain. / indicates that the collaboration method is not compatible with this model/data setting (e.g., model merging requires models to share the same architecture and only works with model pool 1). Results show that diverse model collaboration approaches improve over individual models in 61.0\% (model, evaluation) settings, with \textcolor{Dandelion}{text-level} and \textcolor{OliveGreen}{weight-level} collaboration algorithms as generally stronger.}
\label{tab:big}
\resizebox{1\linewidth}{!}{ 

\begin{tabular}{lccccccc|ccccccc}\toprule[1.5pt]
&\multicolumn{7}{c}{Model Pool 1: Specialized LMs} &\multicolumn{7}{c}{Model Pool 2: General-Purpose LMs} \\\cmidrule{2-15}
&QA &math &reason &safety &code &IF &avg. &QA &math &reason &safety &code &IF &avg. \\\midrule
Best Single &0.588 &0.774 &0.418 &0.586 &0.553 &1.387 &0.549 &0.504 &0.834 &0.413 &0.495 &0.614 &8.402 &0.582 \\ \midrule[0.75pt]
\textcolor{NavyBlue}{\textsc{Cascade}} &\cellcolor{orange!15}0.597 &\cellcolor{orange!15}0.805 &0.387 &\cellcolor{orange!15}0.631 &\cellcolor{orange!15}0.597 &1.385 &\cellcolor{orange!15}0.565 &\cellcolor{orange!15}0.540 &0.831 &\cellcolor{orange!15}0.423 &0.457 &\cellcolor{orange!15}0.632 &\cellcolor{orange!15}9.289 &\cellcolor{orange!15}0.591 \\
\textcolor{NavyBlue}{\textsc{Graph Routing}} &\cellcolor{orange!15}0.610 &\cellcolor{orange!15}0.780 &0.387 &\cellcolor{orange!15}0.638 &\cellcolor{orange!15}0.579 &\cellcolor{orange!15}1.676 &\cellcolor{orange!15}0.565 &\cellcolor{orange!15}0.592 &\cellcolor{orange!15}0.836 &\cellcolor{orange!15}0.487 &\cellcolor{orange!15}0.567 &\cellcolor{orange!15}\underline{0.790} &7.516 &\cellcolor{orange!15}0.644 \\
\textcolor{NavyBlue}{\textsc{Prompt Routing}} &\cellcolor{orange!15}0.596 &\cellcolor{orange!15}0.803 &0.411 &0.571 &\cellcolor{orange!15}0.597 &1.381 &\cellcolor{orange!15}0.558 &\cellcolor{orange!15}0.579 &\cellcolor{orange!15}0.837 &\cellcolor{orange!15}0.461 &\cellcolor{orange!15}0.557 &\cellcolor{orange!15}0.754 &\cellcolor{orange!15}10.276 &\cellcolor{orange!15}0.649 \\
\textcolor{NavyBlue}{\textsc{Switch Generation}} &0.587 &0.733 &0.382 &\cellcolor{orange!15}0.652 &0.412 &-1.060 &0.493 &0.491 &\cellcolor{orange!15}0.860 &\cellcolor{orange!15}0.481 &\cellcolor{orange!15}0.521 &0.395 &\cellcolor{orange!15}\textbf{17.234} &\cellcolor{orange!15}0.625 \\
\textcolor{NavyBlue}{\textsc{Trained Router}} &\cellcolor{orange!15}0.589 &\cellcolor{orange!15}0.802 &\cellcolor{orange!15}0.424 &0.547 &\cellcolor{orange!15}0.588 &1.268 &\cellcolor{orange!15}0.552 &0.490 &0.815 &0.320 &0.474 &0.544 &7.552 &0.539 \\
\textcolor{NavyBlue}{\textsc{Mentor Collab}} &0.524 &0.599 &0.238 &0.526 &0.009 &-3.552 &0.316 &0.396 &0.748 &\cellcolor{orange!15}0.416 &\cellcolor{orange!15}0.543 &0.412 &-6.609 &0.419 \\
\textcolor{NavyBlue}{\textsc{Nudging}} &\cellcolor{orange!15}0.595 &\cellcolor{orange!15}0.789 &0.395 &0.583 &\cellcolor{orange!15}0.596 &0.994 &\cellcolor{orange!15}0.550 &\cellcolor{orange!15}0.545 &0.814 &\cellcolor{orange!15}0.481 &\cellcolor{orange!15}0.508 &\cellcolor{orange!15}0.763 &\cellcolor{orange!15}8.640 &\cellcolor{orange!15}0.625 \\ \midrule[0.75pt]
\textcolor{Dandelion}{\textsc{Heterogeneous Swarms}} &\cellcolor{orange!15}0.633 &\cellcolor{orange!15}0.801 &\cellcolor{orange!15}0.448 &\cellcolor{orange!15}0.587 &\cellcolor{orange!15}0.614 &0.331 &\cellcolor{orange!15}0.563 &\cellcolor{orange!15}0.610 &\cellcolor{orange!15}\underline{0.884} &\cellcolor{orange!15}\underline{0.528} &\cellcolor{orange!15}0.537 &\cellcolor{orange!15}0.728 &\cellcolor{orange!15}9.764 &\cellcolor{orange!15}0.662 \\
\textcolor{Dandelion}{\textsc{Knowledge Card}} &0.494 &0.761 &\cellcolor{orange!15}0.438 &0.488 &0.386 &-0.157 &0.471 &0.492 &0.814 &\cellcolor{orange!15}0.454 &0.438 &\cellcolor{orange!15}0.640 &2.100 &0.534 \\
\textcolor{Dandelion}{\textsc{LLM Blender}} &\cellcolor{orange!15}\textbf{0.657} &\cellcolor{orange!15}\textbf{0.859} &\cellcolor{orange!15}0.450 &0.558 &\cellcolor{orange!15}0.605 &-1.267 &\cellcolor{orange!15}0.550 &\cellcolor{orange!15}\underline{0.635} &\cellcolor{orange!15}0.867 &\cellcolor{orange!15}\textbf{0.530} &0.478 &\cellcolor{orange!15}0.772 &\cellcolor{orange!15}\underline{14.375} &\cellcolor{orange!15}\textbf{0.694} \\
\textcolor{Dandelion}{\textsc{Majority Vote}} &\cellcolor{orange!15}0.622 &0.608 &0.382 &\cellcolor{orange!15}0.601* &/ &/ &\cellcolor{orange!15}0.553 &\cellcolor{orange!15}0.528 &0.658 &0.411 &\cellcolor{orange!15}\underline{0.637}* &/ &/ &0.558 \\
\textcolor{Dandelion}{\textsc{Multiagent Feedback}} &0.496 &0.696 &\cellcolor{orange!15}0.423 &0.419 &0.439 &-3.298 &0.415 &0.470 &0.725 &0.386 &0.429 &\cellcolor{orange!15}0.658 &-2.284 &0.475 \\
\textcolor{Dandelion}{\textsc{Multiagent Finetuning}} &\cellcolor{orange!15}\underline{0.653} &0.739 &\cellcolor{orange!15}0.431 &\cellcolor{orange!15}0.703* &/ &/ &\cellcolor{orange!15}0.631 &\cellcolor{orange!15}0.567 &\cellcolor{orange!15}0.882 &\cellcolor{orange!15}0.503 &\cellcolor{orange!15}\textbf{0.679}* &/ &/ &\cellcolor{orange!15}0.658 \\
\textcolor{Dandelion}{\textsc{Multiagent Refine}} &0.553 &\cellcolor{orange!15}0.816 &\cellcolor{orange!15}0.424 &0.448 &0.553 &-2.946 &0.473 &\cellcolor{orange!15}0.516 &0.799 &\cellcolor{orange!15}0.480 &0.471 &\cellcolor{orange!15}0.649 &1.224 &0.541 \\
\textcolor{Dandelion}{\textsc{Structure}} &0.571 &0.763 &0.372 &\cellcolor{orange!15}0.590 &0.526 &-5.289 &0.448 &\cellcolor{orange!15}0.602 &\cellcolor{orange!15}0.841 &\cellcolor{orange!15}0.496 &0.452 &\cellcolor{orange!15}0.711 &-3.163 &0.541 \\
\textcolor{Dandelion}{\textsc{Agg-LM}} &\cellcolor{orange!15}0.648 &0.743 &0.376 &\cellcolor{orange!15}0.624* &/ &/ &\cellcolor{orange!15}0.598 &\cellcolor{orange!15}\textbf{0.692} &\cellcolor{orange!15}\textbf{0.892} &\cellcolor{orange!15}0.524 &\cellcolor{orange!15}0.635* &/ &/ &\cellcolor{orange!15}\underline{0.686} \\
\textcolor{Dandelion}{\textsc{Sparta}} &0.562 &\cellcolor{orange!15}\textbf{0.859} &\cellcolor{orange!15}\underline{0.478} &0.551 &\cellcolor{orange!15}\textbf{0.789} &\cellcolor{orange!15}\textbf{9.664} &\cellcolor{orange!15}\underline{0.707} &\cellcolor{orange!15}0.590 &\cellcolor{orange!15}0.843 &\cellcolor{orange!15}0.469 &\cellcolor{orange!15}0.556 &\cellcolor{orange!15}\textbf{0.798} &\cellcolor{orange!15}9.883 &\cellcolor{orange!15}0.658 \\ \midrule[0.75pt]
\textcolor{Maroon}{\textsc{Logit Fusion}} &0.499 &0.587 &0.350 &0.563 &0.482 &-0.621 &0.450 & / & / & / & / & / & / & / \\
\textcolor{Maroon}{\textsc{Logit Contrastive}} &0.557 &0.442 &0.291 &\cellcolor{orange!15}0.602 &0.114 &-0.369 &0.374 & / & / & / & / & / & / & / \\ \midrule[0.75pt]
\textcolor{OliveGreen}{\textsc{Dare Ties}} &\cellcolor{orange!15}0.625 &\cellcolor{orange!15}0.814 &\cellcolor{orange!15}0.438 &\cellcolor{orange!15}0.608 &\cellcolor{orange!15}0.675 &\cellcolor{orange!15}1.848 &\cellcolor{orange!15}0.595 & / & / & / & / & / & / & / \\
\textcolor{OliveGreen}{\textsc{Greedy Soup}} &\cellcolor{orange!15}0.621 &\cellcolor{orange!15}0.795 &\cellcolor{orange!15}0.420 &\cellcolor{orange!15}0.701 &\cellcolor{orange!15}0.675 &\cellcolor{orange!15}1.822 &\cellcolor{orange!15}0.603 & / & / & / & / & / & / & / \\
\textcolor{OliveGreen}{\textsc{LoraHub}} &0.454 &\cellcolor{orange!15}0.844 &0.409 &0.575 &0.395 &-0.691 &0.482 & / & / & / & / & / & / & / \\
\textcolor{OliveGreen}{\textsc{Model Swarms}} &\cellcolor{orange!15}0.628 &\cellcolor{orange!15}\underline{0.853} &\cellcolor{orange!15}\textbf{0.497} &\cellcolor{orange!15}\textbf{0.724} &\cellcolor{orange!15}\underline{0.763} &\cellcolor{orange!15}\underline{8.493} &\cellcolor{orange!15}\textbf{0.729} & / & / & / & / & / & / & / \\
\textcolor{OliveGreen}{\textsc{Weight Expo}} &\cellcolor{orange!15}0.604 &\cellcolor{orange!15}0.775 &0.391 &\cellcolor{orange!15}\underline{0.708} &\cellcolor{orange!15}0.728 &1.180 &\cellcolor{orange!15}0.594 & / & / & / & / & / & / & / \\
\bottomrule[1.5pt]
\end{tabular}
\vspace*{-10pt}
}
\end{table*}

\section{Experiment Settings}

\paragraph{Models and Implementation} We employ the two most representative model pool settings, specialized and general-purpose, to fairly benchmark diverse model collaboration approaches. Model pool \#1 features 3 specialized LLMs \citep{jiang2025sparta} fine-tuned on different domains of data in Tulu-v3 \citep{lambert2024tulu}; model pool \#2 features 3 general-purpose LLMs, specifically \textsc{Qwen-2.5-7B} \citep{qwen2}, \textsc{Llama-3.1-8B} \citep{grattafiori2024llama}, and \textsc{Olmo-3-7B} \citep{olmo2025olmo}. Note that some model collaboration algorithms might be designed for other specific settings (e.g., switch generation for pretrained-aligned collaboration \citep{feng2025don}): we encourage users to experiment with diverse model settings and choose collabroation strategies based on their need. We by default employ 512 max new tokens, 1024 for code generation, temperature $\tau = 0.7$, and top-p $p=0.9$ sampling for text generation. We employ the default hyperparameters provided in \ourmethod{} for diverse model collaboration approaches.

\paragraph{Data and Evaluation} We evaluate collaboration strategies across six domains of 11 datasets: QA (AGIeval, MMLU-redux), math (MATH, GSM8k), reasoning (BigBench-hard, TheoremQA), safety (CocoNot, TruthfulQA), coding (HumanEval), and instruction following (Alpaca, Human Interest). We employ task accuracy, generative verifiers \citep{ma2025general}, and reward models \citep{liu2024skywork} included in \ourmethod{} to evaluate their corresponding tasks, and report the macro-average of tasks within a domain.

\section{Results}
\label{sec:results}
Table \ref{tab:big} shows the performance of model collaboration approaches on two model pools and 6 evaluation domains.

\paragraph{Model collaboration is broadly effective.} We highlight improvements over initial models without collaboration in \colorbox{orange!15}{orange} cells: diverse model collaboration strategies bring performance gains across six evaluation domains in 61.0\% of settings, demonstrating their general effectiveness across varying models and tasks.

\begin{figure*}
    \centering
    \includegraphics[width=1\linewidth]{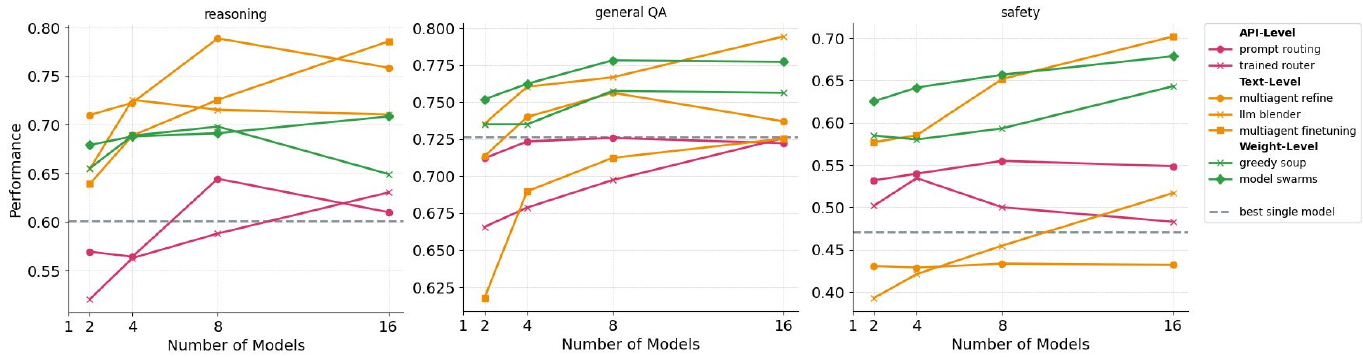}
    \caption{Scaling the number of models in model collaboration systems and evaluating on reasoning, QA, and safety domains. We observe a consistent upward trend that further improves over the best single model, with text-level and weight-level methods being more scalable and benefiting from a larger pool of diverse models. This indicates that by scaling up model collaboration, we could build bottom-up compositional AI systems where the components are small but the system is large.}
    \vspace*{-10pt}
    \label{fig:scaling_number}
\end{figure*}

\begin{figure}[t]
\vskip 0.2in
\begin{center}
\centerline{\includegraphics[width=1\linewidth]{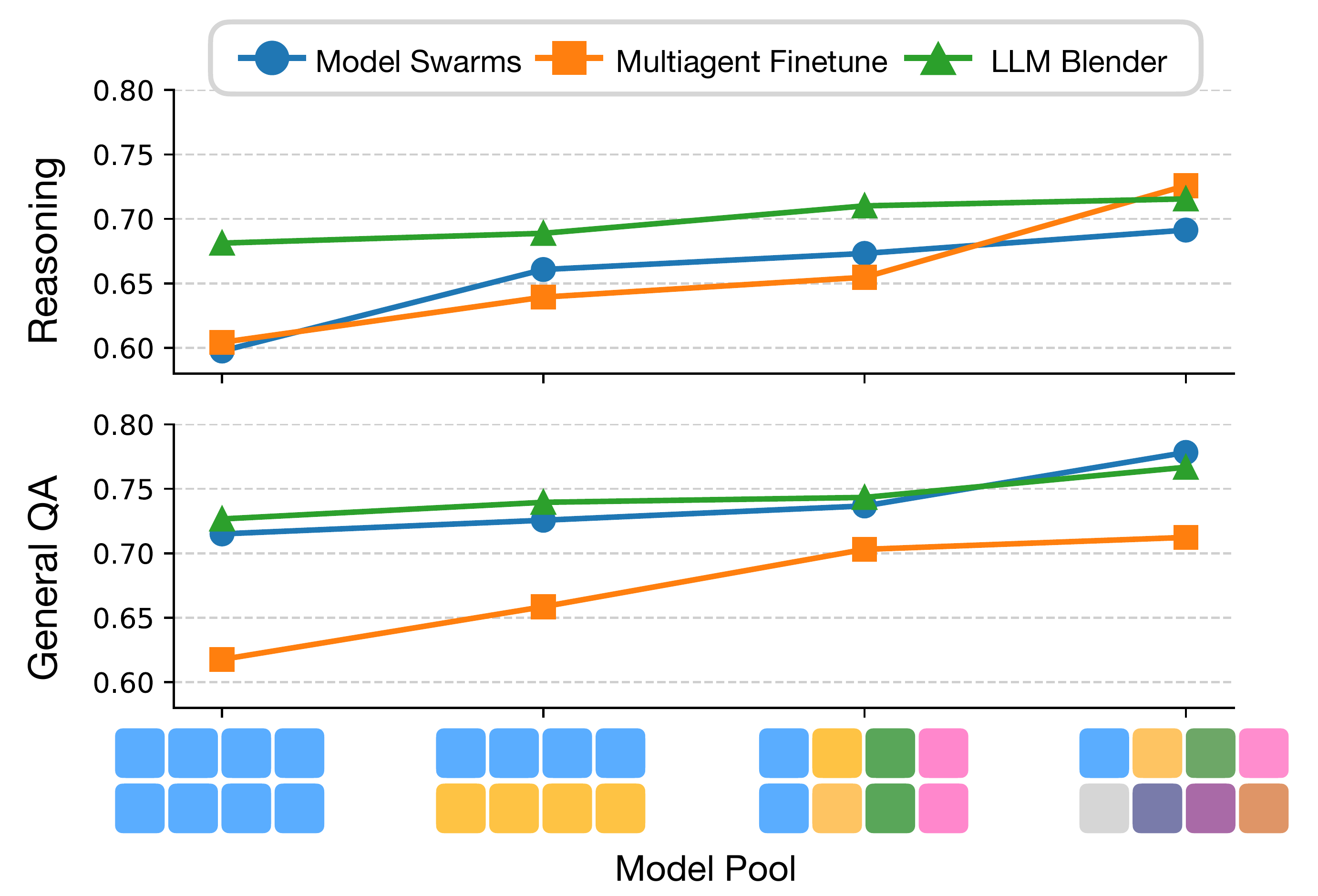}}
\caption{Impact of model pool diversity on collaboration performance. The x-axis shows the configurations of model pool diversity: $1\times 8, 2 \times 4, 4 \times 2$ and $8 \times 1$. Results demonstrate that model collaboration benefits from increased diversity among participating models, indicating the need for model specialization.}
\label{fig:diversity}
\end{center}
\vskip -0.2in
\end{figure}

\paragraph{\ourmethod{} offers fair comparison and new insights about diverse collaboration strategies.} \textcolor{OliveGreen}{Weight-level} is in general the most effective, achieving an average performance of 60.1 compared to the global average of 53.5. These approaches operate with assumptions that the participating models share the same architecture, in order to perform merging or arithmetic directly with model parameters. Among all approaches, \textcolor{OliveGreen}{Model Swarms}, \textcolor{Dandelion}{Sparta Alignment},  \textcolor{Dandelion}{LLM Blender}, and \textcolor{Dandelion}{Agg-LM} are among the best with three of them being \textcolor{Dandelion}{text-level} collaboration methods. This indicates that collaboration by models exchanging generated texts with each other is both broadly applicable and strong.

\paragraph{\ourmethod{} reveals the synergy among collaboration strategies, application domains, and model settings.} \textcolor{NavyBlue}{Trained router} works better with specialized LMs (pool \#1) than general LMs (pool \#2), potentially due to the artificial hivemind phenomenon \citep{jiangartificial} where general-purpose LMs generate similar responses to certain types of queries, reducing the effectiveness of routing. \textcolor{Dandelion}{Multiagent Refine} works great with math and reasoning tasks, but refining generation is challenging for safety and refusal scenarios in the CocoNot dataset. Approaches such as \textcolor{Dandelion}{Sparta Alignment} and \textcolor{OliveGreen}{Model Swarms} bring consistent improvements to almost all model and dataset settings, thanks to their methodology of tailoring models and collaboration to diverse tasks and applications. By running, evaluating, and comparing diverse model collaboration algorithms with \ourmethod{}, we could derive a treasure trove of insights to reflect on existing methods and motivate future work.

\section{Analysis}

\begin{figure*}
    \centering
    \includegraphics[width=0.9\linewidth]{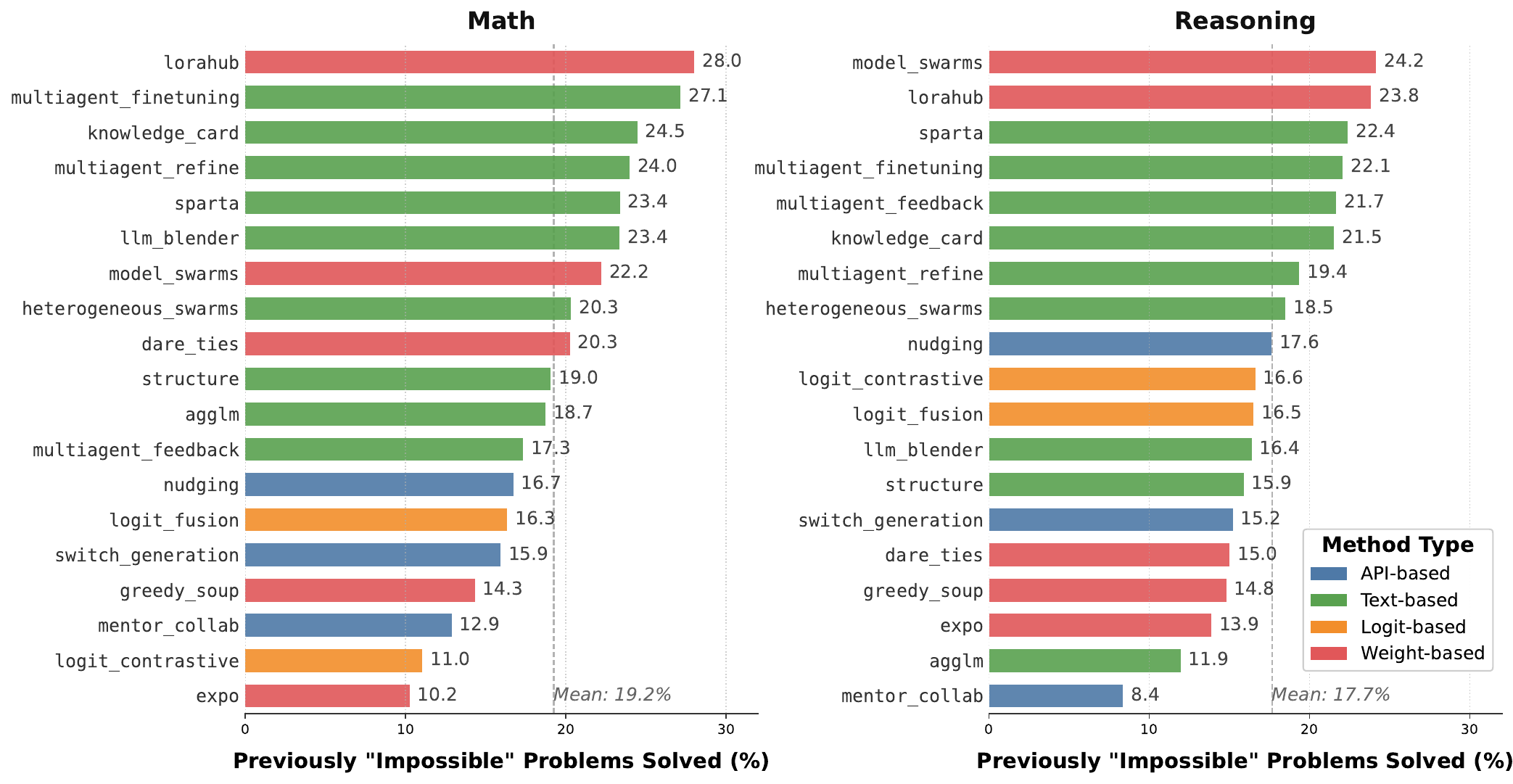}
    \caption{For problems where none of the LLMs could solve individually, what percentage of them are solvable with the model collaboration system, across diverse tasks and collaboration strategies. We observe consistent \emph{collaborative emergence} across settings with an average of 18.5\%, indicating that many model collaboration algorithms do not merely offer a union of existing capabilities: new skills emerge in the collaborative system of multiple models that solve problems where individual models struggle to.}
    \label{fig:impossible}
\end{figure*}

\paragraph{Scaling the number of models} We experiment with scaling up model collaboration systems by scaling the number of participating models in collaboration algorithms. We experiment with 2, 4, 8, and 16 LLMs sourced from academic research artifacts (details in Appendix \ref{app:experiment_details}) and evaluate with strong methods across collaboration levels. Figure \ref{fig:scaling_number} demonstrates that model collaboration algorithms are generally scalable: from 2 models to 16 models, there are consistent upward trends across reasoning, QA, and safety evaluation domains. Among them, text-level and weight-level methods are more successful than API-level routing approaches: scaling the number of models increases the candidate pool of routing and might introduce noise, while having more models collaborate via generated texts or model parameters offer deeper integration and stronger synergy.

\begin{figure}[t]
    \centering
    \includegraphics[width=0.8\linewidth]{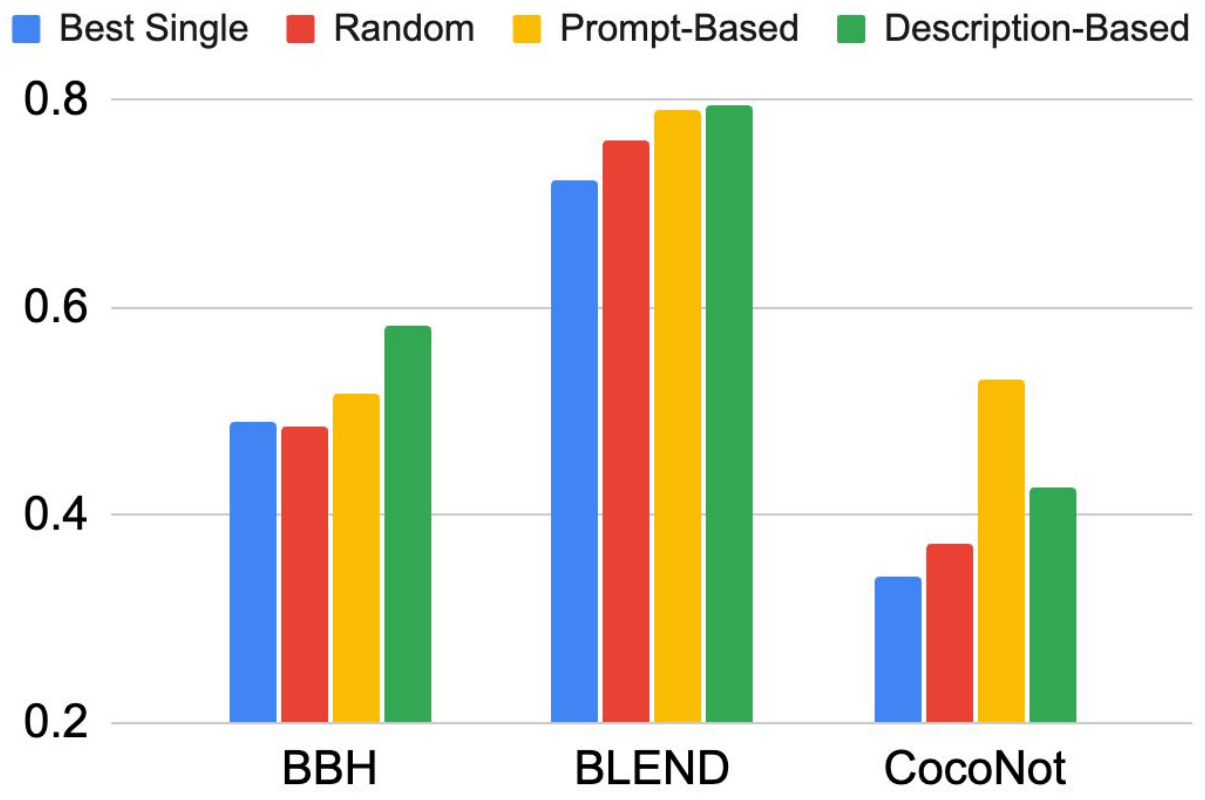}
    \caption{Employing random, prompt-based, or description-based strategies to select 3 models out of 8 for collaboration. Both strategies outperform the random baseline and no collaboration, indicating the importance of model selection strategies and highlighting the need for future research.}
    \vspace*{-10pt}
    \label{fig:select}
\end{figure}

\paragraph{Scaling the diversity of models} We posit that the benefit of model collaboration not only comes from more compute, but also from the diversity and complementary strength of multiple LLMs. We experiment with $a \times b$ settings: $a$ unique LLMs each repeated for $b$ times, so that the total pool size is fixed as $ab$. We experiment with $1 \times 8$, $2 \times 4$, $4 \times 2$, and $8 \times 1$ settings and present results in Figure \ref{fig:diversity}, showing a consistent upward trend with the increase of model diversity. This suggests that employing a pool of diverse language models is key to the success of model collaboration, and motivates future research on studying model diversity, training diverse language models, and more.

\paragraph{Collaborative emergence} For a pool of multiple LLMs, the most complex and difficult problems might be beyond their capability, and none could individually solve them. However, we observe that model collaboration systems built with these LLMs could sometimes solve these problems ``impossible'' for the individual models, a phenomenon we term \emph{collaborative emergence}. We quantify what percentage of these previously ``impossible'' problems are now solvable with the collaborative system with model pool \#1 across diverse tasks and collaboration strategies. Figure \ref{fig:impossible} demonstrates that collaborative emergence is a consistent phenomenon across various settings, with an average of 18.5\% problems now solvable with model collaboration algorithms. We present additional collaborative emergence results on more evaluation domains in Appendix \ref{app:additional_analysis}.

\paragraph{How to select models} In section \ref{sec:results}, we experiment with two of the most popular model collaboration settings: a pool of specialized LLMs fine-tuned on different domains of data, and a pool of general-purpose LLMs released by different entities. However, for diverse applications, how to dynamically select models that offer a diverse set of related expertise for collaboration remains an open research question. \ourmethod{} empowers this investigation: we take an initial step with two model selection strategies. 1) \emph{prompt-based selection}, where an LLM (\textsc{Qwen-2.5-7B}) is given the task description, the description of all candidate LLMs, and asked to select a subset of them for collaboration. 2) \emph{similarity-based selection}, employing an encoder LM (\textsc{RoBERTa-base}) to encode model descriptions, calculate pairwise distances of description embeddings, and select a subset with the most intra-group distances. We experiment with the first 8 models in Figure \ref{fig:scaling_number} and select 3, comparing with baselines of the best single model and average performance over 5 random selections. Figure \ref{fig:select} shows that both strategies outperform random selection and no collaboration. We envision future research on dynamic model selection in model collaboration systems uniquely supported by \ourmethod{}.

\paragraph{Discussion} \ourmethod{} offers flexible implementations of diverse model collaboration algorithms, uniquely enabling the study and investigation of a wide spectrum of research questions about model collaboration, compositional AI systems, collaborative development, and more.

\begin{itemize}[leftmargin=*]
    \item How do we scale up model collaboration? Which collaboration strategies are more scalable to large quantities of diverse LLMs, so that we could build compositional AI systems where the modular components are small but the system is large?
    \item How do we achieve bottom-up collaborative development for AI systems? Specifically, how do models trained by different stakeholders together form decentralized systems where no one has unilateral control over state-of-the-art AI?
    \item How does the cost/efficiency of diverse collaboration strategies compare?\footnotemark[3]\footnotetext[3]{We present an analysis of existing algorithms of \ourmethod{} in Appendix \ref{app:additional_analysis}.} How do we improve efficiency and design novel and cost-effective collaboration algorithms (e.g., through information exchange at the latent space \citep{wu2025improved})?
    \item What are the risks of having malicious models in model collaboration systems? How do we safeguard decentralized collaborative AI systems from malicious actors and artifacts?
    \item How do we train models that are not only individually strong, but also \emph{compositionally strong}: models that bring new information, improve on underepresented skills, and boost existing models when used in collaboration?
\end{itemize}

We envision the creation of \ourmethod{} as turbocharging future research on these and many other important topics for an open, compositional, and decentralized AI future.

\section{Related Work}

Section \ref{sec:method} provides a comprehensive overview of model collaboration methods and algorithms. In addition to individual methods, here we focus on high-level related works from both conceptual and engineering standpoints.

\citet{du2024position} puts forward the position that ``a single is not all you need'' and advocates for compositional generative systems,  discussing their benefits across vision, reinforcement learning, robotics, and a brief mention of language. \citet{feng2025one} later presents a deepdive on model collaboration specifically to language and language models, characterizing diverse collaboration strategies based on four levels of information exchange across models. \citet{raffel2023building} proposes to ``build machine learning models like open source software'' and advocates for community-based training and updating of AI models: the community and decentralized aspects are important points underpinning the need for model collaboration. Many survey papers also summarize collaborative and modular approaches towards building AI models and systems \citep{yadavsurvey, cai2025survey, wang2025modular}.

From an engineering standpoint, \ourmethod{} is related to a few resources for modular and collaborative AI. MergeKit \citep{goddard-etal-2024-arcees} features implementations of diverse model merging approaches and is widely employed: for some of the weight-level collaboration algorithms in \ourmethod{} we employ MergeKit, and we gratefully acknowledge their contribution. LLMRouter \citep{llmrouter2025} and RouteLLM \citep{ongroutellm} are open-source libraries that support diverse methods for routing queries among multiple LLMs, overlapping with some of the API-level methods. In contrast, \ourmethod{} focuses on the whole spectrum of model collaboration algorithms, from API-level routing and switching, to text-level collaboration where models exchange generated texts, to logit-level arithematic on the token probabilities of multiple LMs, to weight-level merging and parameter-space methods. We encourage readers to check out the related valuable resources that inspired \ourmethod{}.

\vspace*{5pt}
\section{Conclusion}
We present \ourmethod{}, a comprehensive toolkit and resource for model collaboration research. \ourmethod{} integrates 26 model collaboration algorithms spanning four levels of cross-LLM information exchange, 25 evaluation datasets spanning diverse application domains, and is seamlessly extensible for new novel methods and evaluation datasets. Extensive experiments with \ourmethod{} demonstrate that model collaboration approaches boost participating LLMs in 61.0\% of cases, showing gains across a wide range of domains such as reasoning, safety, coding, and more. Further analysis showcases the scalability of model collaboration, the important benefits of model diversity, and highlights collaborative emergence --- how model collaboration systems solve challenging problems where the individual models can not. We envision \ourmethod{} as uniquely enabling and empowering novel and diverse research questions about model collaboration, compositional AI, and collaborative development.

\section*{Impact Statement}
As an open resource to facilitate model collaboration research, it is possible that malicious actors attempting to influence AI models/systems with a certain agenda would also investigate how a malicious model/component could impact/jailbreak compositional AI systems. As such, we envision important future work on the safety of model collaboration systems: studying the impact of malicious models in decentralized model collaboration systems, designing strategies to identify and mitigate their impact, and more. \ourmethod{} empowers this endeavor: by allowing the stress-testing and red-teaming of compositional AI systems and implementing diverse guardrail strategies, we will be ready to defend the integrity of future decentralized AI systems from the outset.

\bibliography{example_paper}
\bibliographystyle{icml2026}

\newpage
\appendix
\onecolumn

\section{Additional Analysis}
\label{app:additional_analysis}

\paragraph{Training/Inference Complexity} We analyze the training and inference time complexity of different collaboration methods in Table~\ref{tab:complexity}. Most collaboration methods require an additional training stage to determine the collaboration structure. Moreover, collaboration at different levels incurs varying computational costs, with text-level methods exhibiting relatively higher time complexity.

\begin{wraptable}{r}{0.4\textwidth}
\centering
\scriptsize
\setlength{\tabcolsep}{3pt}
\renewcommand{\arraystretch}{1}
\resizebox{1\linewidth}{!}{
\begin{tabular}{lccc}\toprule[1.5pt]
&mmlu-redux &gsm8k &bbh \\\midrule[0.75pt]
w/o model 1 &0.572 &0.729 &0.539 \\
w/o model 2 &0.581 &0.759 &0.561 \\
w/o model 3 &0.589 &0.822 &0.565 \\
w/o model 4 &0.593 &0.828 &0.588 \\
w/o model 5 &0.598 &0.872 &0.598 \\
avg &0.587 &0.802 &0.570 \\
std &0.010 &0.057 &0.023 \\
\bottomrule[1.5pt]
\end{tabular}
}
\vspace*{10pt}
\caption{Leave-one-out analysis to study the sensitivity of model collaboration, specifically the multiagent debate approach, to minor changes in model composition. An average standard deviation of 0.030 shows that it is mostly robust.}
\label{tab:sensitivity}
\end{wraptable}

\begin{figure*}[t]
  \centering 
  \includegraphics[width=0.6\textwidth]{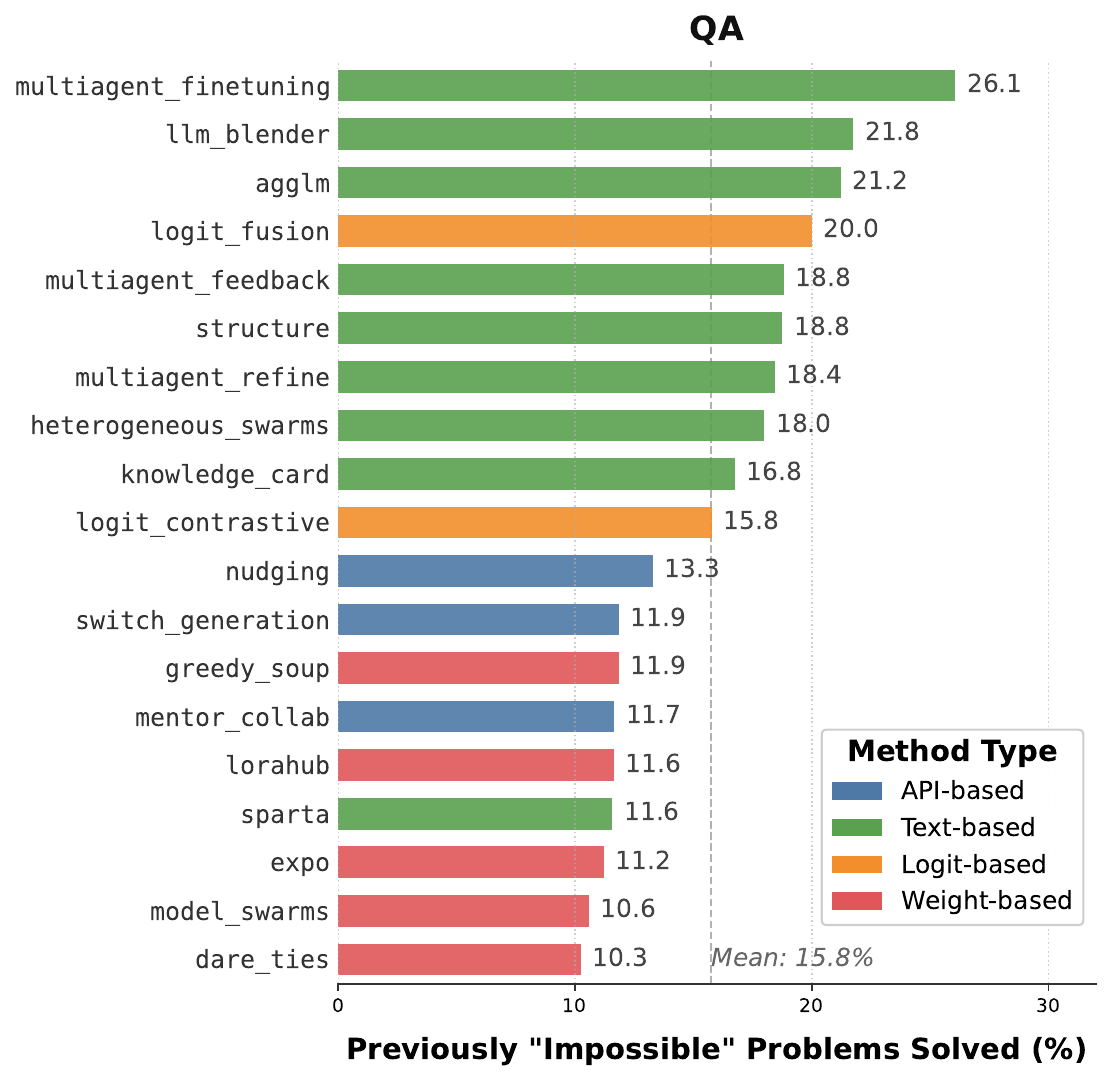}
  \vspace{-0.1in}
  \caption{
    Collaborative emergence on the General-purpose QA domain.
  }
  \label{analysis_impossible_qa}
\end{figure*}

\begin{figure*}[t]
  \centering 
  \includegraphics[width=0.6\textwidth]{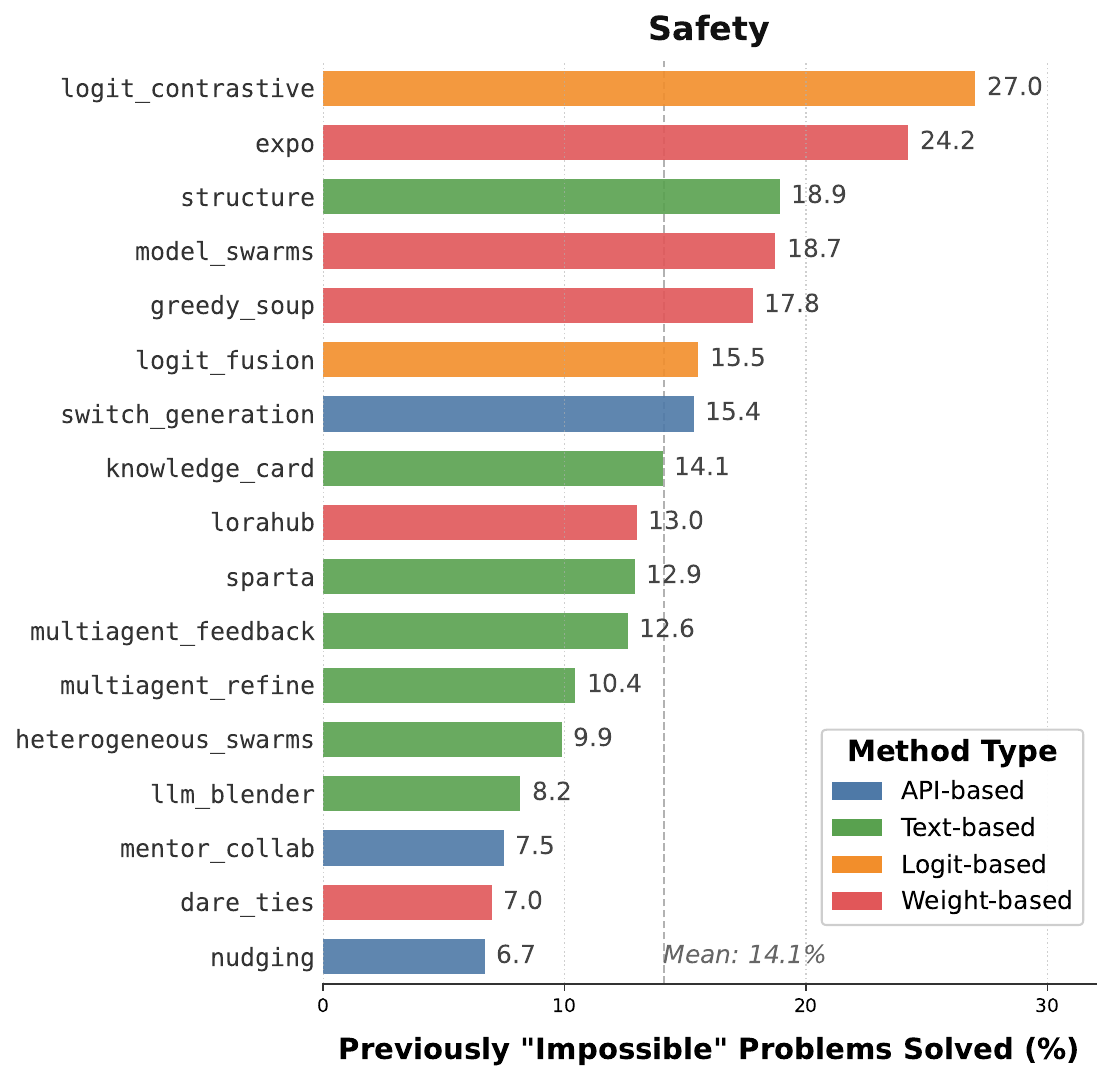}
  \vspace{-0.1in}
  \caption{
    Collaborative emergence on the Safety domain.
  }
  \label{analysis_impossible_safety}
\end{figure*}

\begin{figure*}[t]
  \centering 
  \includegraphics[width=0.6\textwidth]{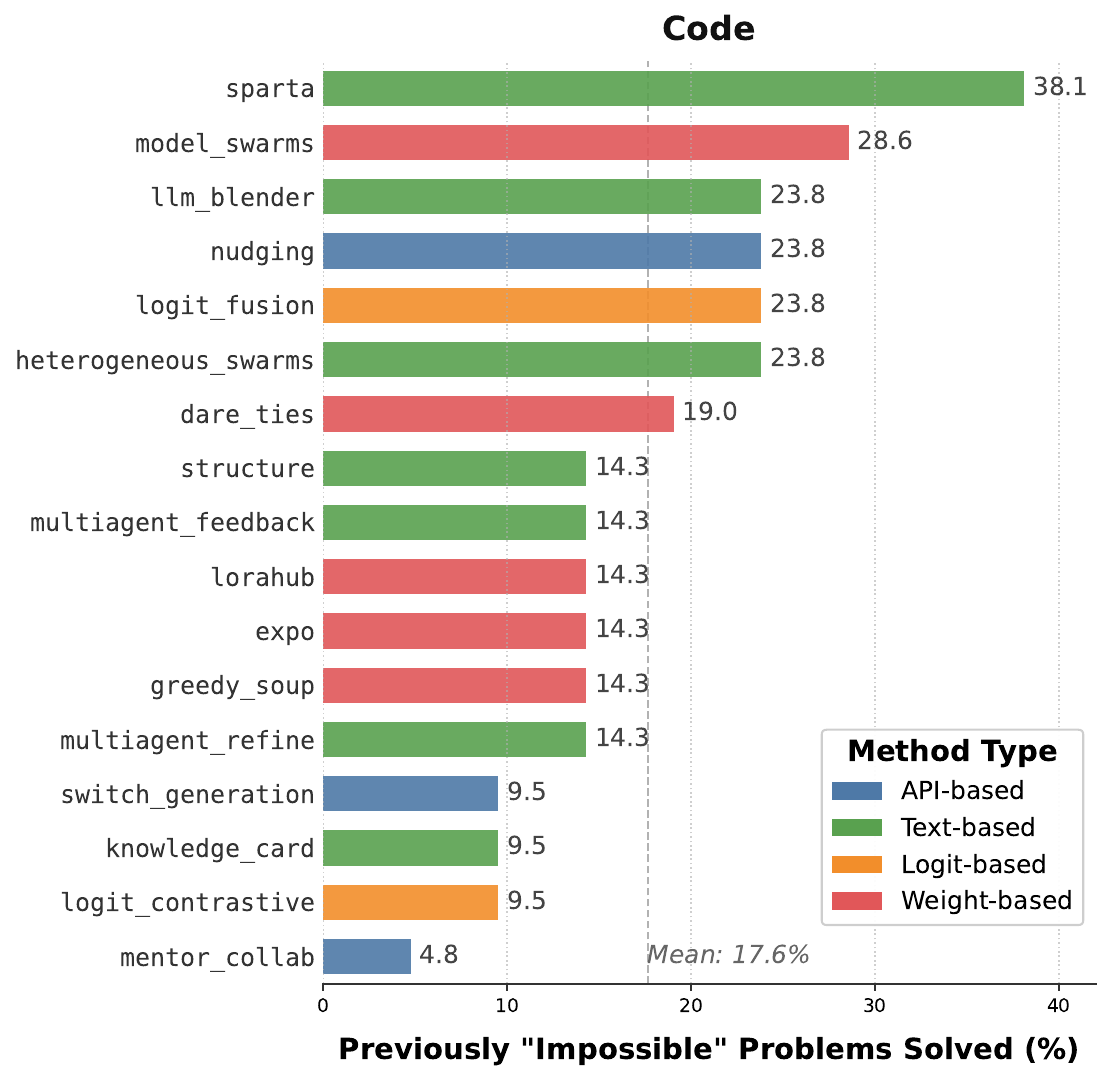}
  \vspace{-0.1in}
  \caption{
    Collaborative emergence on the Coding domain.
  }
  \label{analysis_impossible_code}
\end{figure*}

\paragraph{Collaborative Emergence on More Domains} We extend our analysis of collaborative emergence to additional evaluation domains: General-purpose QA, Safety, and Coding. Figures \ref{analysis_impossible_qa}, \ref{analysis_impossible_safety}, and \ref{analysis_impossible_code} present the percentage of previously ``impossible'' problems—those unsolvable by any individual LLM—that become solvable through model collaboration.
For the General-purpose QA domain, we observe a mean collaborative emergence rate of 15.8\%, with multiagent\_finetuning (26.1\%) and llm\_blender (21.8\%) achieving the highest rates. In the Safety domain, logit\_contrastive achieves the highest emergence rate of 27.0\%, followed by expo at 24.2\%, resulting in a mean of 14.1\%. The Coding domain exhibits a strong collaborative emergence with a mean of 17.6\%, where sparta alignment achieves a remarkable 38.1\% and model swarms follows at 28.6\%. These results reinforce that collaborative emergence is a robust phenomenon across diverse tasks, and different collaboration strategies exhibit varying strengths depending on the application domain.

\paragraph{Sensitivity to Model Choice} Ideally, model collaboration strategies should be robust to minor changes in participating models. We create a testbed of model choice sensitivity by employing the first 5 models in Figure \ref{fig:scaling_number}, and conduct leave-one-out analysis with the multiagent debate strategy across three datasets. Results in Table \ref{tab:sensitivity} demonstrate that it is mostly robust with an average standard deviation of 0.03 across five model pool settings, with math and reasoning being more sensitive and would thus benefit from tailored model selection strategies.

\section{Experiment Details}
\label{app:experiment_details}

\paragraph{Dataset Details} 

We systematically report datasets details integrated in \ourmethod{}, including reasoning, math, QA, knowledge, instruction following, safety, coding and social science. \ourmethod{} is easily extensible and users can easily incorporate additional datasets in \ourmethod{}. Datasets statistics are summarized in Table \ref{tab:dataset_statistics}.

\begin{table*}[t]
\centering
\scriptsize
\setlength{\tabcolsep}{3pt}
\renewcommand{\arraystretch}{1}
\resizebox{0.5\linewidth}{!}{
\begin{tabular}{lccc}
\toprule[1.5pt]
\multirow{2}{*}{Dataset} &\multirow{2}{*}{Source} &\multicolumn{2}{c}{Size} \\\cmidrule{3-4}
& &dev &test \\\midrule
AGIeval & \citep{zhong2024agieval} & 1156 & 1156 \\
Alpaca & \citep{dubois2023alpacafarm} & 2000 & 1000 \\
ARC-Challenge & \citep{clark2018think} & 299 & 1172 \\
BigBench-hard & \citep{suzgun2023challenging} & 1000 & 1000 \\
BLEND & \citep{myung2024blend} & 1000 & 1000 \\
CocoNot & \citep{brahman2024art} & 1000 & 1000 \\
CultureBench & \citep{chiu2024culturalbench} & 2454 & 2454\\
GPQA (Diamond) & \citep{rein2024gpqa} & 99 & 99 \\
GPQA (Extended) & \citep{rein2024gpqa} & 273 & 273 \\
GPQA (Main) & \citep{rein2024gpqa} & 224 & 224 \\
GSM8k & \citep{cobbe2021training} & 200 & 1000 \\
Human Interest & \citep{feng2025model} & 400 & 400 \\
HumanEval & \citep{chen2021codex} & 50 & 114 \\
MATH & \citep{hendrycks2021measuring} & 956 & 956 \\
MBPP & \citep{austin2021program} & 487 & 487 \\
MedMCQA &\citep{pal2022medmcqa} & 2091 & 2092 \\
MedQA & \citep{jin2021disease} & 636 & 637 \\
MMLU-redux & \citep{gema2025we} & 1000 & 1000 \\
PopQA & \citep{mallen2023not} & 1000 & 1000 \\
PubMedQA & \citep{jin2019pubmedqa} & 500 & 500 \\
ScienceMeter & \citep{wang2025sciencemeter} & 1000 & 1000 \\
SciRIFF & \citep{wadden2025sciriff} & 114 & 139 \\
TableMWP & \citep{ludynamic} & 375 & 376 \\
TableMWP (MC) & \citep{ludynamic} & 124 & 125 \\
TheoremQA & \citep{chen2023theoremqa} & 400 & 400 \\
TruthfulQA & \citep{lin2022truthfulqa} & 200 & 617 \\
WikiDyK & \citep{zhang2025bidirectional} & 1000 & 765\\
WildChat & \citep{zhaowildchat} & 1000 & 1000 \\
\bottomrule[1.5pt]
\end{tabular}
}
\vspace*{10pt}
\caption{Datasets Statistics.}
\label{tab:dataset_statistics}
\end{table*}

\paragraph{Implementation Details} We by default employ 512 max new tokens, with an exception of 1024 for coding tasks, and $\tau = 0.7$ and $p = 0.9$ for temperature and top-p sampling in text generation. We employ the default hyperparameters in \ourmethod{} for different model collaboration algorithms. Model pool \#1 includes the following three models from \citep{jiang2025sparta}: \textsc{bunsenfeng/yuru\_qw\_wizardlm}, \textsc{bunsenfeng/yuru\_qw\_sharegpt}, \textsc{bunsenfeng/yuru\_qw\_oasst1}. Model pool \#2 includes the following three models: \textsc{Qwen/Qwen2.5-7B-Instruct}, \textsc{meta-llama/Llama-3.1-8B-Instruct}, \textsc{allenai/Olmo-3-7B-Instruct}. We evaluate instruction following in Table \ref{tab:big} with the Skywork reward model (\textsc{Skywork/Skywork-Reward-Llama-3.1-8B-v0.2}).

For Figure \ref{fig:scaling_number}, we employ 16 LLMs in the following order: \textsc{chtmp223/Qwen2.5-7B-CLIPPER} \citep{pham2025clipper}, \textsc{chengq9/ToolRL-Qwen2.5-3B} \citep{qian2025toolrl}, \textsc{AgentFlow/agentflow-planner-7b} \citep{li2025flow}, \textsc{nanami/ladder-last16L-llama3.1-8binstruct-sft4k-stage2v03-bsize32-rkl8b} \citep{zhangladder}, \textsc{viswavi/qwen2.5\_rlcf} \citep{viswanathanchecklists}, \textsc{milli19/promptmii-llama-3.1-8b-instruct} \citep{xiao2025prompt}, \textsc{Zhengping/conditional-probability-regression} \citep{wang2025always}, \textsc{yale-nlp/MDCure-Qwen2-7B-Instruct} \citep{liu2025mdcure}, \textsc{GritLM/GritLM-7B} \citep{muennighoff2024generative}, \textsc{lime-nlp/Qwen2.5-7B-Instruct-SUM10} \citep{song2025hallucination}, \textsc{geyang627/care-chinese-gemma2-9b} \citep{guo2025care}, \textsc{bespokelabs/Bespoke-Stratos-7B} \citep{tang2024minicheck}, \textsc{kangdawei/Llama-3.1-8B-Instruct-GenderNeutral-Finetuned} \citep{wei2025mitigating}, \textsc{DeepRetrieval/DeepRetrieval-PubMed-3B-Llama} \citep{jiang2025deepretrieval}, \textsc{yale-nlp/MDCure-Qwen2-1.5B-Instruct} \citep{liu2025mdcure}, \textsc{Zhaoxuan/PUGC-Mistral-DPO} \citep{tan2025aligning}. To run weight-level approaches with these models, we perform distillation with these models as teacher and \textsc{Qwen-2.5-7B} as student using Tulu-v3 data \citep{lambert2024tulu} to standardize model architecture.

\paragraph{Releasing \ourmethod{}} \ourmethod{} is publicly available at \href{https://github.com/BunsenFeng/model_collaboration}{https://github.com/BunsenFeng/model\_collaboration}. We will also release a PyPI package based on \ourmethod{} for command-line execution. We commit to offer continuous support for \ourmethod{} even after publication: working with external contributors to add their collaboration algorithms, add new datasets, update PyPI package versions, and more.

\begin{table*}[t]
\vskip 0.15in
\setlength{\tabcolsep}{3pt}
\renewcommand{\arraystretch}{1.2}
\begin{center}
\begin{small}
\begin{sc}
\resizebox{1\linewidth}{!}{
\begin{tabular}{lccc}
\toprule[1.5pt]
Method    & Training     & Inference    & Notes \\
\midrule[0.75pt]
Cascade   & / & $ 2  D m \cdot \sum_{i=1}^{n} \frac{k_i}{2^{i-1}}$ & \textnormal{Each level defer 50\%} \\
Graph Router & $2 n D m \cdot \max(k_i)$ & $2  D m \cdot \max(k_i)$ & \textnormal{Omit GNN training} \\
Prompt Router & / & $2 D m \cdot (k_r + \max(k_i))$ & \textnormal{Router $k_r$}\\
LLM Router & $2nD m \cdot \max(k_i)+6rDM \cdot k_r$ & $2  D  m \cdot (k_r+\max(k_i))$ & \textnormal{Router $k_r$} \\
Switch Generate  & $6 r D  m  \cdot k_s$ & $2 D  m \cdot \bigl(k_s/\textit{patch} + \max(k_i)\bigr)$ & \textnormal{Switcher $k_s$, rounds $r$, path size $\textit{patch}$}\\
Mentor Collab & $2n D  m \cdot \max(k_i)$ & $2 D m \cdot \mathrm{mean}(k_i)$ & \textnormal{Omit MLP training} \\
Co-llm & $2nD m \cdot \mathrm{max}(k_i)$ & $2D m \cdot \mathrm{mean}(k_i)$& / \\
Nudging & /& $2 D m \cdot \max(k_i)$ &  / \\
\midrule[0.75pt]
Heterogeneous Swarms & $2 n D m \cdot \max(k_i)$ & $2 Gr D m \cdot \max(k_i)$ & \textnormal{Graph generation $G$, rounds $r$}\\
Knowledge Card      & $2 n D m \cdot \max(k_i)$ & $2nD m \cdot \max(k_i)$ & / \\
LLM Blender & $2nD m\cdot\max(k_i)+6 rDM \cdot (n^2k_r+ k_f)$ & $2nDM \cdot \max(k_i)+2DM\cdot (k_r+k_f)$ & \textnormal{ranker $k_r$, fuser $k_f$, rounds $r$}        \\
Majority Vote   & /& $2nD m \cdot \max(k_i)$ &  / \\
Multiagent Refine & $2nD m  \cdot\max(k_i)$ & $2nrD m \cdot\max(k_i)$& \textnormal{rounds $r$} \\
Multiagent Feedback & $2nDm \cdot \max(k_i)$ & $2nrfD m \cdot \max(k_i)$ & \textnormal{Feedback $f$, rounds $r$}\\
Multiagent Finetuning & $2nD m\cdot \max(k_i)+6nrDM \cdot 2\max(k_i)$ & $2nrD m \cdot \max(k_i)$ & \textnormal{rounds $r$} \\
Structure & $2nD m\cdot\max(k_i)$ & $2GrD m \cdot\max(k_i)$ & \textnormal{structure $G$, rounds $r$}\\
Agg-LM & $2nsDM\cdot \max(k_i)+6rDM \cdot \min(k_i)$ & $2D m \cdot (n\max(k_i)+\min(k_i))$ & \textnormal{sample size $s$, rounds $r$} \\
Sparta & $2nD m \cdot \max(k_i)+6nrDM \cdot \max(k_i)$ & $2nD m \cdot \max(k_i)$ & \textnormal{training rounds $r$}\\
\midrule[0.75pt]
Logit Fusion &  / & $2nD m \cdot\max(k_i)$ & / \\
Logit Contrastive & $2nD m  \cdot\max(k_i)$ & $2nD m \cdot\max(k_i)$ & / \\
\midrule[0.75pt]
Dare Ties & / & $2D m\cdot\max(k_i)$ & / \\
Greedy Soup & $2(2n-1)D m  \cdot\max(k_i)$ & $2D m\cdot\max(k_i)$ & \\
LoraHub & $2nD m\cdot\max(k_i)$ & $2D m \cdot\max(k_i)$ & / \\
Model Swarms & $2nrD m  \cdot\max(k_i)$ & $2D m  \cdot\max(k_i)$ & \textnormal{training rounds $r$}\\
Weight ExPO & $2nD m \cdot\max(k_i)$ & $2 D m \cdot\max(k_i)$ & / \\

\bottomrule[1.5pt]
\end{tabular}
}
\end{sc}
\end{small}
\end{center}
\vspace*{10pt}
\caption{Collaboration Methods: Training and Inference FLOPs Complexity Analysis. $D$ denotes the dataset size, $m$ the maximum tokens length and the model pool $\mathcal{K}=\{k_i\}^n_1$ contains $n$ models. Each forward pass produces 2 FLOPs per parameter, and each backward pass produces 6 FLOPs per parameter.}
\label{tab:complexity}
\end{table*}

\end{document}